%% file: root.tex
\documentclass[letterpaper, 10 pt, conference]{ieeeconf}  

\IEEEoverridecommandlockouts                              

\usepackage{graphics} 
\usepackage{epsfig} 
\usepackage{amsmath} 
\usepackage{amssymb}  
\usepackage{tikz}
\usepackage{float}

\usepackage{booktabs}
\usepackage{algorithm}
\usepackage{algpseudocode}
\usepackage{multirow}
\usepackage{hyperref}
\newtheorem{theorem}{Theorem}

\DeclareMathOperator*{\argmax}{argmax}
\DeclareMathOperator*{\argmin}{argmin}

\title{\LARGE \bf VICAN: Very Efficient Calibration Algorithm for Large Camera Networks}

\author{Gabriel Moreira$^{1,2}$  Manuel Marques$^{2}$  João Paulo Costeira$^{2}$ and Alexander Hauptmann$^{1}$%
\thanks{$^{1}$Language Technologies Institute, Carnegie Mellon University, School of Computer Science, USA
        {\tt\small $\{$gmoreira, alex$\}$@cs.cmu.edu}.}%
\thanks{$^{2}$Institute for Systems and Robotics, Instituto Superior Técnico,
        Lisboa, Portugal
        {\tt\small $\{$manuel, jpc$\}$@isr.tecnico.ulisboa.pt}.}%
}

\makeatletter
\let\@oldmaketitle\@maketitle
\renewcommand{\@maketitle}{\@oldmaketitle
  \includegraphics[width=\linewidth]
    {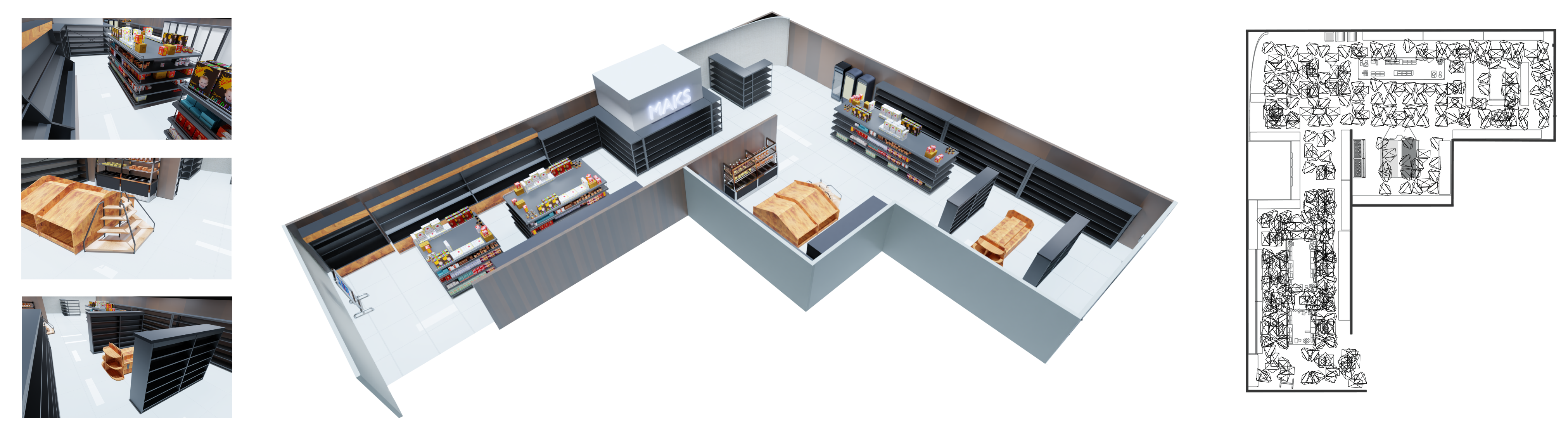}\par\centering\small  \refstepcounter{figure}Fig.~\thefigure. Our method is motivated by applications in smart-retail that require fast pose estimation of large camera networks. The image in the center shows a 358m$^2$ store on which 342 cameras were mounted according to the locations and orientations depicted on the right. From over 100000 images of a moving object, as the three examples on the left illustrate, our algorithm computes accurate pose estimates in a short timespan.\setcounter{figure}{1}}
\makeatother

\begin{document}
\maketitle

\thispagestyle{empty}
\pagestyle{empty}

\begin{abstract}
The precise estimation of camera poses within large camera networks is a foundational problem in computer vision and robotics, with broad applications spanning autonomous navigation, surveillance, and augmented reality. In this paper, we introduce a novel methodology that extends state-of-the-art Pose Graph Optimization (PGO) techniques. Departing from the conventional PGO paradigm, which primarily relies on camera-camera edges, our approach centers on the introduction of a dynamic element - any rigid object free to move in the scene - whose pose can be reliably inferred from a single image. Specifically, we consider the bipartite graph encompassing cameras, object poses evolving dynamically, and camera-object relative transformations at each time step. This shift not only offers a solution to the challenges encountered in directly estimating relative poses between cameras, particularly in adverse environments, but also leverages the inclusion of numerous object poses to ameliorate and integrate errors, resulting in accurate camera pose estimates. Though our framework retains compatibility with traditional PGO solvers, its efficacy benefits from a custom-tailored optimization scheme. To this end, we introduce an iterative primal-dual algorithm, capable of handling large graphs. Empirical benchmarks, conducted on a new dataset of simulated indoor environments, substantiate the efficacy and efficiency of our approach.
\end{abstract}

\section{INTRODUCTION}
In this paper, we address the problem of localizing a network of calibrated cameras distributed in 3D space. The importance of this task lies in its role as a prerequisite: accurately determining the pose of each camera within a shared reference frame is paramount for enabling spatial awareness, in order to facilitate applications ranging from augmented reality and tracking to action recognition. Without precise camera pose estimates, the capabilities of camera networks are significantly limited.

Modern approaches to solve the camera network localization problem fall in one of two groups. Structure-from-Motion (SfM) systems, such as COLMAP \cite{Schonberger2016Structure-from-MotionRevisited}, simultaneously reconstruct the 3D scene and solve for the camera poses. In large scenes however, this can be prohibitive from a computational standpoint. On the other hand, Pose Graph Optimization (PGO) approaches \cite{Kummerle2011G2o:Optimization, CarloneLagrangianSolutions,Carlone2015InitializationOptimization, Dellaert2012FactorIntroduction, Martinec2007RobustReconstruction, Dellaert2012FactorIntroduction}, also referred to as motion synchronization \cite{Arrigoni2016SpectralSE3, Rosen2016SE-Sync:Group} or averaging, \cite{Hartley2013RotationAveraging, AlvaroParra2021RotationAveraging} and consensus algorithms \cite{Tron2009DistributedNetworks, Tron2014DistributedMeasurements}, provide an expedited two-step way of solving the problem. Initially, each image is probed for keypoints, which are then matched with those extracted from other images. For each image pair with enough point correspondences, the relative direction and rotation between the respective cameras is estimated via the epipolar constraint. Subsequently, a non-convex optimization problem is solved in order to localize the cameras in a common reference frame. Its optimum is the maximum likelihood point estimate (MLE) of the set of camera poses. 

In recent years, the MLE formulation of motion synchronization relying on the bi-dual of the original non-convex problem, a Semidefinite Program (SDP), has seen wide adoption in the literature pertaining to Visual Simultaneous Localization in Mapping (SLAM) \cite{CarloneLagrangianSolutions, Rosen2016SE-Sync:Group} and rotation synchronization \cite{Eriksson2018RotationDuality, DellaertShonanN, Moreira2021RotationGraphs, AlvaroParra2021RotationAveraging}. Under the moderate noise levels commonly found in most applications, this SDP relaxation is tight, yielding the optimum of the PGO problem. Fast and non-iterative spectral approximations for the camera rotations have also been proposed \cite{Arrigoni2016SpectralSE3, Moreira2021FastFactorization}, yielding estimates surprisingly close to the optimum \cite{Doherty2022PerformanceSLAM}. 

Despite these notable achievements, for the same camera network topology, the synchronization step is only as good as the pairwise transformations \cite{Doherty2022PerformanceSLAM,Rosen2016SE-Sync:Group}. In low-light, low-texture, or occluded regions of the environment, the acquisition of precise point correspondences becomes challenging. This subsequently hinders the accurate estimation of relative camera poses, ultimately restricting the efficacy of PGO. These difficulties are further exacerbated in poorly connected, or `bottleneck', regions of the graph, which obstruct the redistribution of the pairwise errors \cite{Boumal2012Cramer-RaoRotations}.

As a means to circumvent the aforementioned issues, we introduce a novel methodology. Alongside the static camera network, we incorporate a known dynamic rigid object, within the field-of-view of a subset of the cameras, such that its pose can be reliably inferred from a single image. Our focal point is the bipartite pose graph comprising the ensemble of static camera poses, the object poses evolving over time, and the set of relative transformations binding them together. The indirect estimation of camera poses through the interplay of camera-object and pairwise object transformations allows us to mitigate some of the adverse aspects of camera pose estimation, by integrating a large number of camera-object  transformations.

While our novel problem formulation initially appears amenable to conventional motion synchronization solvers such as g2o \cite{Kummerle2011G2o:Optimization}, GTSAM \cite{Dellaert2012FactorIntroduction}, SE-Sync \cite{Rosen2016SE-Sync:Group} and MAKS \cite{Moreira2021FastFactorization}, which exhibit a node-agnostic nature regarding graph elements, it is important to note that these methods were originally tailored for graphs composed exclusively of camera nodes. Within the PGO literature, which predominantly targets SLAM applications, datasets frequently encompass tens of thousands of pairwise transformations. In our framework, the introduction of moving objects adds a multitude of new nodes and edges with each time step. Consequently, existing solvers prove inadequate in handling these augmented graphs. To this end, we propose a fast iterative primal-dual algorithm to solve for the cameras and object rotations. 

In summary, we make the following contributions
\begin{itemize}
    \item We propose a new MLE formulation of the camera network localization problem via a bipartite camera-object pose graph.
    \item We introduce a new iterative method capable of handling a large number of object nodes from image streams. When compared to ground-truth poses, our method achieves average rotation and translation errors of 0.04deg and 3cm in a 358m$^2$ shop covered by 342 cameras (Fig. 1), and of 0.07deg and 0.7cm in a 72m$^2$ room covered by 25 cameras (Fig. \ref{fig:small_room}).
\end{itemize}

The remainder of the paper\footnote{Code, dataset and extended version of the paper with full derivations are available in \url{https://github.com/gabmoreira/vican}.} is organized as follows. We introduce our formulation of the problem and the iterative scheme to solve it in Section \ref{sec:proposed_method}. In Section \ref{sec:experiments} we present a new image and 3D dataset for camera network pose estimation and benchmarks of our method therein. Concluding remarks are drawn in Section \ref{sec:conclusions}.

\section{PROPOSED METHOD}
\label{sec:proposed_method}

\begin{figure}
    \centering
    \input{fig/measurements.tikz}
    \caption{Standard PGO (left) vs our augmentation with object nodes (right). Pairwise relative transformations are shown as $\tilde{\mathbf{P}}_{\cdot,\cdot}$. The $i$-th camera node is $c_i$, and $m_i^{(t)}$ is the $i$-th object node at time $t$.}
    \label{fig:pgo_vs_augmented}
\end{figure}
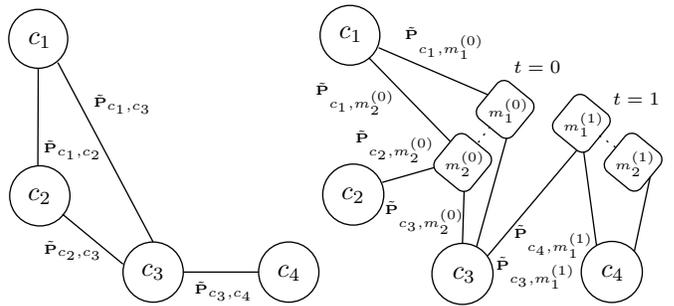

Consider a static network of calibrated cameras, indexed by $c\in\mathcal{C}$, with $C=|\mathcal{C}|$. The pose of the $c$-th camera w.r.t. an arbitrary but fixed reference frame is given by $\mathbf{P}_c \in \mathrm{SE}(3) \subset \mathbb{R}^{4\times 4}$, or equivalently $(\mathbf{R}_c, \mathbf{t}_c) \in \mathrm{E}(3)$. We then consider a rigid object that is free to move in $\mathbb{R}^3$. We decompose this object into $M$ nodes which may correspond \textit{e.g.}, to $M$ fiducial markers, indexed via $m\in\mathcal{M}$. We indicate the pose of the $m$-th object node at time $t$, for $t\in\mathcal{T}$, as $\mathbf{P}_{m^{(t)}}$. Our goal is to estimate $\{c\in\mathcal{C} : \mathbf{P}_c \}$ using pairwise camera-object relative transformations $\tilde{\mathbf{P}}_{c,m^{(t)}}$, obtained \textit{e.g.}, via Perspective-n-Point (PnP). We assume the following convention for the composition of transformations $\mathbf{P}_{ij}=\mathbf{P}_i \mathbf{P}_j^{-1}$.

Denote by $G=(\mathcal{V},\mathcal{E})$, the bipartite graph with $\mathcal{V}=\mathcal{C}\cup (\mathcal{T}\times \mathcal{M})$ \textit{i.e.}, each camera and each object node at each time instant are associated with a vertex. The latter component of the graph is assumed to be much larger than the former. The edge $(c,m^{(t)})\in\mathcal{E}$ corresponds to an image captured by camera $c$ at time $t$ of the $m$-th object node. The relative pose measurement of this node w.r.t. the camera is denoted as $(\tilde{\mathbf{t}}_{c,m^{(t)}}, \tilde{\mathbf{R}}_{c,m^{(t)}})$. A comparison between traditional PGO and our formulation is showcased in Fig. \ref{fig:pgo_vs_augmented}.

\subsection{Maximum likelihood estimation}
Under the assumption of isotropic Gaussian noise with precision $\tau_{c,m^{(t)}}$ in the translation variables and isotropic Langevin noise \cite{Boumal2012Cramer-RaoRotations} with concentration $k_{c,m^{(t)}}$ in the rotations, the negative log-likelihood function (NLL) \cite{CarloneLagrangianSolutions} reads as
\begin{align}
    &-\log f = \sum_{c,m^{(t)}\in\mathcal{E}} \frac{\tau_{c,m^{(t)}}}{2}\|\tilde{\mathbf{t}}_{c,m^{(t)}}-\mathbf{t}_c +   \nonumber \\ 
    &\mathbf{R}_{c}\mathbf{R}_{m^{(t)}}^\top \mathbf{t}_{m^{(t)}} \|_2^2 -k_{c,m^{(t)}}\mathrm{Tr}\left(\tilde{\mathbf{R}}_{c,m^{(t)}}\mathbf{R}_{m^{(t)}} \mathbf{R}_c^\top\right).
    \label{eq:pgo_neg_log}
\end{align}
Instead of optimizing jointly the poses of the $C$ cameras and the $M\times T$ object nodes, we rely on the rigidity of the object and further assume that we have access to the pairwise relative transformation between any two object nodes at the same time instant, $m_i^{(t)}$ and $m_j^{(t)}$, denoted $\bar{\mathbf{P}}_{m_i^{(t)},m_j^{(t)}}$. 
We will henceforth drop the time superscript and write $\bar{\mathbf{P}}_{m_i,m_j}=(\bar{\mathbf{R}}_{m_i, m_j}, \bar{\mathbf{t}}_{m_i, m_j})$. Without loss of generality, we have thus,  
\begin{equation}
\forall t\in\mathcal{T}\; \forall m\in\mathcal{M}\;
\begin{cases}
     \mathbf{R}_{m^{(t)}} = \bar{\mathbf{R}}_{m,m_1} \mathbf{R}_{m_1^{(t)}} \\
     \mathbf{t}_{m^{(t)}} = \bar{\mathbf{R}}_{m,m_1} \mathbf{t}_{m_1^{(t)}} + \bar{\mathbf{t}}_{m, m_1}
     \label{eq:change_of_variables}
\end{cases}
\end{equation}
\textit{i.e.}, the pose of each marker at time $t$ is given relatively to the pose of the object in that instant, $\mathbf{P}_{m_1^{(t)}}$. Thus, we only optimize over $\{t\in\mathcal{T}: \mathbf{P}_{m_1^{(t)}}\}$. 
Introducing (\ref{eq:change_of_variables}) in the first term of the NLL (\ref{eq:pgo_neg_log}) yields
\begin{align}
    \sum_{t}\sum_{c,m^{(t)}\in\mathcal{E}}\frac{\tau_{c,m^{(t)}}}{2}\bigg\|\tilde{\mathbf{t}}_{c,m^{(t)}} + \mathbf{R}_c \mathbf{R}_{m^{(t)}}^\top \bar{\mathbf{t}}_{m,m_1} \nonumber \\
    -\left(\mathbf{t}_c - \mathbf{R}_{c}\mathbf{R}_{m^{(t)}}^\top \bar{\mathbf{R}}_{m,m_1} \mathbf{t}_{m_1^{(t)}}\right) \bigg\|_2^2.
    \label{eq:translations_term}
\end{align}
The second term becomes
\begin{align}
    \sum_{t\in\mathcal{T}c\in\mathcal{C}} - \mathrm{Tr}\left(\sum_{m^{(t)}\in N(c)}\left(k_{c,m^{(t)}}\tilde{\mathbf{R}}_{c,m^{(t)}}\bar{\mathbf{R}}_{m,m_1}\right) \mathbf{R}_{m_1^{(t)}} \mathbf{R}_c^\top\right),
    \label{eq:rotations_term}
\end{align}
where $N(c)$ denotes the neighborhood of the $c$-th camera, \textit{i.e.}, the set of markers that are visible from it. In summary, we simplify the graph by merging all the objects nodes at each time instant $t$, into a single object node at time $t$, corresponding to the marker $m_{1}$. 

We will now turn our attention to the minimization of the non-convex NLL with terms (\ref{eq:translations_term}) and (\ref{eq:rotations_term}). To this end we will adopt the strategy used in \cite{Martinec2007RobustReconstruction,Moreira2021FastFactorization}, wherein the problem is decoupled in: non-convex rotations synchronization by minimizing (\ref{eq:rotations_term}) and a least-squares problem that yields the translations by minimizing (\ref{eq:translations_term}). The main difficulty of this method lies in optimally and efficiently solving the former.

\subsection{Synchronization of rotations} 
Define the block-matrix $\tilde{\mathbf{R}}_{\mathcal{C}\mathcal{T}}\in\mathbb{R}^{3C\times 3T}$, indexed via $c$ and $t$, as
\begin{equation}
    [\tilde{\mathbf{R}}_{\mathcal{C}\mathcal{T}}]_{c,t} := 
        \sum_{m :\; m^{(t)}\in N(c)}k_{c,m^{(t)}}\tilde{\mathbf{R}}_{c,m^{(t)}}\bar{\mathbf{R}}_{m,m_1}.
    \label{eq:merged_object_rtilde}
\end{equation}
We assume $k_{c,m^{(t)}} = 0$ if $(c,m^{(t)})\not\in\mathcal{E}$. The block-entry $c,t$ of $\tilde{\mathbf{R}}_{\mathcal{C}\mathcal{T}}$ is a weighted sum of rotation measurements from camera $c$ to marker $m_1$ at time $t$. Define the pairwise block matrix $\mathbf{\tilde{R}}\in\mathbb{R}^{3(C+T)\times 3(C+T)}$ as
\begin{equation}
    \mathbf{\tilde{R}} := 
    \begin{bmatrix}
        \mathbf{0} & \mathbf{\tilde{R}}_{\mathcal{C}\mathcal{T}} \\
        \mathbf{\tilde{R}}_{\mathcal{C}\mathcal{T}}^\top & \mathbf{0} \\
    \end{bmatrix}
\end{equation}
and the block-vector $\mathbf{R} \in \mathrm{SO}(3)^{3(C+T)}\subset \mathbb{R}^{3(C+T)\times 3}$ as
\begin{align}
    \mathbf{R} &:= 
     \begin{bmatrix}
    \mathbf{R}_{\mathcal{C}}^\top & \mathbf{R}_{\mathcal{T}}^\top
    \end{bmatrix}^\top \nonumber \\
    &=\begin{bmatrix}
    \mathbf{R}_{c_1}^\top & \dots & \mathbf{R}_{c_C}^\top  & \mathbf{R}_{m_1^{(1)}}^\top  & \dots & \mathbf{R}_{m_1^{(T)}}^\top
    \end{bmatrix}^\top.
\end{align}
The set of rotations that minimize (\ref{eq:rotations_term}) is the solution of the rotation synchronization problem
\begin{align}
    \min_{\mathbf{R}\in\mathrm{SO}(3)^{(C+T)}} -\mathrm{Tr}\left(\tilde{\mathbf{R}} \mathbf{R}\mathbf{R}^\top\right).
    \label{eq:rotation_syncronization_problem}
\end{align}
The adjacency matrix associated with this problem, denoted $\mathbf{A} \in \mathbb{R}^{(C+T)\times (C+T)}$, can be written in the same fashion
\begin{equation}
    \mathbf{A} := 
    \begin{bmatrix}
        \mathbf{0} & \mathbf{A}_{\mathcal{C}\mathcal{T}} \\
        \mathbf{A}_{\mathcal{C}\mathcal{T}}^\top & \mathbf{0}
    \end{bmatrix},
\end{equation}
with entry $c,t$ given by
\begin{equation}
    [\mathbf{A}_{\mathcal{C}\mathcal{T}}]_{c,t} := \sum_{m:\;m^{(t)}\in N(c)}k_{c,m^{(t)}}.
    \label{eq:merged_object_adj}
\end{equation}

The KKT and optimality conditions of Problem (\ref{eq:rotation_syncronization_problem}), which can be found in \cite{Eriksson2018RotationDuality}, are as follows. Denote by $\mathbf{\Lambda}\in\mathbb{R}^{3(C+T)\times 3(C+T)}$, the symmetric $3\times 3$ block-diagonal dual variable. The first-order KKT condition reads as $\mathbf{\Lambda}\mathbf{R} = \tilde{\mathbf{R}}\mathbf{R}$. If a primal-dual pair $(\mathbf{R}^\ast,\mathbf{\Lambda}^\ast)$ verifies the KKT, then $\mathbf{\Lambda}^\ast-\tilde{\mathbf{R}}\succeq 0$ is a sufficient optimality condition.

\begin{theorem} Assume strong duality holds and denote by $(\mathbf{\Lambda}^\ast, \mathbf{R}^\ast)$ a primal-dual optimal pair of Problem (\ref{eq:rotation_syncronization_problem}), where the dual variable $\mathbf{\Lambda}$ is decomposed as  
\begin{equation}
    \mathbf{\Lambda} = \begin{bmatrix}
        \mathbf{\Lambda}_\mathcal{C} & \mathbf{0} \\
        \mathbf{0} & \mathbf{\Lambda}_{\mathcal{T}}
    \end{bmatrix},
\end{equation}
with $\mathbf{\Lambda}_{\mathcal{C}}\in\mathbb{R}^{3C\times 3C}$ and $\mathbf{\Lambda}_{\mathcal{T}}\in\mathbb{R}^{3T\times 3T}$. The optimal camera poses $\mathbf{R}_\mathcal{C}^\ast$ are a solution of 
\begin{equation}
    \min_{\mathbf{R}_\mathcal{C}\in\mathrm{SO}(3)^C} -\mathrm{Tr}\left(\left( \tilde{\mathbf{R}}_{\mathcal{C}\mathcal{T}}\mathbf{\Lambda}_{\mathcal{T}}^{\ast^{-1}}\tilde{\mathbf{R}}_{\mathcal{C}\mathcal{T}}^\top\right)\mathbf{R}_\mathcal{C}\mathbf{R}_\mathcal{C}^\top\right)
    \label{eq:new_problem_in_c}
\end{equation}
and the optimal object poses $\mathbf{R}_\mathcal{T}^\ast$ a solution of
\begin{equation}
    \min_{\mathbf{R}_\mathcal{T}\in\mathrm{SO}(3)^T} -\mathrm{Tr}\left(\left( \tilde{\mathbf{R}}_{\mathcal{C}\mathcal{T}}^\top\mathbf{\Lambda}_{\mathcal{C}}^{\ast^{-1}}\tilde{\mathbf{R}}_{\mathcal{C}\mathcal{T}}^\top\right)\mathbf{R}_\mathcal{T}\mathbf{R}_\mathcal{T}^\top\right).
    \label{eq:new_problem_in_t}
\end{equation}
\label{theo:equivalent_problems}
\end{theorem}

This results leads us then to the following block-coordinate descent updates, which produce feasible primal-dual pairs $(\mathbf{R}^{(k)},\mathbf{\Lambda}^{(k)})$ \textit{i.e.}, at the $k$-th iteration $\mathbf{R}^{(k)}\in\mathrm{SO}(3)^{(C+T)}$ and $\mathbf{\Lambda}^{(k)}-\tilde{\mathbf{R}}\succeq 0$, that maximize the dual function \textit{i.e.}, $-\mathrm{Tr}(\mathbf{\Lambda}^{(k)}) \leq -\mathrm{Tr}(\mathbf{\Lambda}^{(k+1)})$,
\begin{align}
        \mathbf{R}^{(k)}_\mathcal{C} &= \argmin_{\mathbf{R}_\mathcal{C}} -\mathrm{Tr}\left(\left( \tilde{\mathbf{R}}_{\mathcal{C}\mathcal{T}}\mathbf{\Lambda}_{\mathcal{T}}^{(k-1)^{-1}}\tilde{\mathbf{R}}_{\mathcal{C}\mathcal{T}}^\top\right)\mathbf{R}_\mathcal{C}\mathbf{R}_\mathcal{C}^\top\right) \label{eq:update_1}\\
        \mathbf{\Lambda}_\mathcal{C}^{(k)} &= \mathrm{blkdiag} \left( \tilde{\mathbf{R}}_{\mathcal{C}\mathcal{T}}\mathbf{\Lambda}_{\mathcal{T}}^{(k-1)^{-1}}\tilde{\mathbf{R}}_{\mathcal{C}\mathcal{T}}^\top \mathbf{R}_{\mathcal{C}}^{(k)} \mathbf{R}_{\mathcal{C}}^{(k)^\top}\right) \label{eq:update_2} \\
        \mathbf{R}^{(k)}_\mathcal{T} &= \argmin_{\mathbf{R}_\mathcal{T}} -\mathrm{Tr}\left(\left( \tilde{\mathbf{R}}_{\mathcal{C}\mathcal{T}}^\top\mathbf{\Lambda}_{\mathcal{C}}^{(k)^{-1}}\tilde{\mathbf{R}}_{\mathcal{C}\mathcal{T}}^\top\right)\mathbf{R}_\mathcal{T}\mathbf{R}_\mathcal{T}^\top\right) \label{eq:update_3} \\
        \mathbf{\Lambda}_\mathcal{T}^{(k)} &= \mathrm{blkdiag} \left( \tilde{\mathbf{R}}_{\mathcal{C}\mathcal{T}}^\top\mathbf{\Lambda}_{\mathcal{C}}^{(k)^{-1}}\tilde{\mathbf{R}}_{\mathcal{C}\mathcal{T}} \mathbf{R}_{\mathcal{T}}^{(k)} \mathbf{R}_{\mathcal{T}}^{(k)^\top}\right) \label{eq:update_4}
\end{align}
We assume that strong duality holds. Then, the update of $\mathbf{\Lambda}_\mathcal{C}^{(k)}$ in (\ref{eq:update_2}) can be written as
\begin{equation}
        \mathbf{\Lambda}_{\mathcal{C}}^{(k)} = \argmax_{\mathbf{\Lambda}:\;\mathbf{\Lambda_\mathcal{C}}-\tilde{\mathbf{R}}_\mathcal{CT}\mathbf{\Lambda}_\mathcal{T}^{(k-1)^{-1}}\tilde{\mathbf{R}}_\mathcal{CT}^\top\succeq 0} -\mathrm{Tr}(\mathbf{\Lambda}_\mathcal{C}).
\end{equation}
From Schur's complement, the positive semidefinite constraint is equivalent to $\mathbf{\Lambda}-\tilde{\mathbf{R}} \succeq 0$ and $\mathbf{\Lambda}_\mathcal{T} = \mathbf{\Lambda}_\mathcal{T}^{(k-1)}$. We thus rewrite the update as
\begin{equation}
    \mathbf{\Lambda}_{\mathcal{C}}^{(k)} = \argmax_{\mathbf{\Lambda}_\mathcal{C}:\;\mathbf{\Lambda}-\tilde{\mathbf{R}}\succeq 0,\;\mathbf{\Lambda}_\mathcal{T}=\mathbf{\Lambda}_\mathcal{T}^{(k-1)}} -\mathrm{Tr}(\mathbf{\Lambda}).
\end{equation}
Same reasoning applies for update (\ref{eq:update_4}), which yields
\begin{equation}
    \mathbf{\Lambda}_{\mathcal{T}}^{(k)} = \argmax_{\mathbf{\Lambda}_\mathcal{T}:\;\mathbf{\Lambda}-\tilde{\mathbf{R}}\succeq 0,\; \mathbf{\Lambda}_\mathcal{C}=\mathbf{\Lambda}_\mathcal{C}^{(k)}} -\mathrm{Tr}(\mathbf{\Lambda}).
\end{equation}
Therefore, the sequence $\{-\mathrm{Tr}(\mathbf{\Lambda}^{(k)})\}_{k\geq 1}$ is dual feasible and non-decreasing, as desired. The caveat with this approach is that, assuming $T >> C$,  (\ref{eq:update_3}) is more expensive to solve than (\ref{eq:update_1}), and thus the gain is minimal. 

We address this by replacing the minimization in (\ref{eq:update_3}) by a step of the Frank-Wolfe method, also known as the generalized power method (GPM) \cite{Boumal2016NonconvexSynchronization}, which has a closed-form solution. We have then
\begin{align}
    &\mathbf{R}^{(k)}_\mathcal{C} = \argmin_{\mathbf{R}_\mathcal{C}} -\mathrm{Tr}\left(\left( \tilde{\mathbf{R}}_{\mathcal{C}\mathcal{T}}\mathbf{\Lambda}_{\mathcal{T}}^{(k-1)^{-1}}\tilde{\mathbf{R}}_{\mathcal{C}\mathcal{T}}^\top\right)\mathbf{R}_\mathcal{C}\mathbf{R}_\mathcal{C}^\top\right) \label{eq:new_update_1}\\
    &\mathbf{\Lambda}_\mathcal{C}^{(k)} = \mathrm{blkdiag} \left( \tilde{\mathbf{R}}_{\mathcal{C}\mathcal{T}}\mathbf{\Lambda}_{\mathcal{T}}^{(k-1)^{-1}}\tilde{\mathbf{R}}_{\mathcal{C}\mathcal{T}}^\top \mathbf{R}_{\mathcal{C}}^{(k)} \mathbf{R}_{\mathcal{C}}^{(k)^\top}\right) \label{eq:new_update_2} \\
    &\mathbf{U}_t\mathbf{\Sigma}_t\mathbf{V}^\top_t \stackrel{\mathrm{SVD}}= \sum_c\tilde{\mathbf{R}}_{\mathcal{C}\mathcal{T}_{c,t}}^\top\mathbf{R}_{\mathcal{C}_c}^{(k)},\; t\in\mathcal{T} \\
    &\mathbf{\Lambda}_{\mathcal{T}_t}^{(k)} = \mathbf{U}_t\mathbf{\Sigma}_t\mathbf{U}_t^\top,\; t\in\mathcal{T}.
    \label{eq:new_update_3}
\end{align}
To see why this is a reasonable approximation note that, given $\mathbf{R}_\mathcal{C}^{(k)}$, the Frank-Wolfe, or GPM iteration is
\begin{equation}
    \mathbf{R}_\mathcal{T}^{(k)} = \argmin_{\mathbf{R}_\mathcal{T}\in\mathrm{SO}(3)^T}-\mathrm{Tr}\left(\tilde{\mathbf{R}}_\mathcal{CT}\mathbf{R}_\mathcal{C}^{(k)}\mathbf{R}_\mathcal{T}^\top\right),
\end{equation}
which has a closed form solution given by the orthogonal projection $\mathbf{R}_{\mathcal{T}_t}^{(k)} = \mathbf{U}_t\mathbf{V}_t^\top$. Close to the optimum, we must have $\mathbf{R}_\mathcal{C}^{(k)}\approx 
\mathbf{\Lambda}_{\mathcal{C}}^{(k)^{-1}}\tilde{\mathbf{R}}_{\mathcal{C}\mathcal{T}}^\top\mathbf{R}_\mathcal{T}^{(k)}$, from the KKT condition. Hence, $\mathbf{\Lambda}_{\mathcal{T}_t}^{(k)}\approx \mathbf{U}_t\mathbf{\Sigma}_t\mathbf{U}_t^\top$, with equality at the optimum.

In order to set $\mathbf{\Lambda}^{(0)}_\mathcal{T}$, we leverage the spectral initialization \cite{Arrigoni2016SpectralSE3,Moreira2021FastFactorization}, whose distance to the optimum is dependent on the connectivity of the graph and the noise level \cite{Doherty2022PerformanceSLAM}. Given the diagonal graph degree matrix $\mathbf{\Delta}=\mathrm{Diag}(\mathbf{\Delta}_\mathcal{C}, \mathbf{\Delta}_\mathcal{T})$, where
\begin{equation}
    \mathbf{\Delta}_{i,i} = \sum_{j} \mathbf{A}_{i,j},
\end{equation}
this approximation consists of projecting the eigenspace of $\mathbf{\Delta} \otimes \mathbf{I}_3 - \tilde{\mathbf{R}}$ corresponding to the three smallest eigenvalues, to $\mathrm{SO}(3)$. From the KKT condition, \cite{Moreira2021RotationGraphs} interpreted the degree matrix as a dual variable approximation, where $\mathbf{\Lambda}^\ast \approx \mathbf{\Delta} \otimes \mathbf{I}_3$. It can be shown that $\mathbf{\Delta} \otimes \mathbf{I}_3$ is, in fact, dual feasible. Thus, we set $\mathbf{\Lambda}_\mathcal{T}^{(0)}$ to $\mathbf{\Delta}_\mathcal{T} \otimes \mathbf{I}_3$.

In this approach, the bulk of the optimization is carried out in update (\ref{eq:new_update_1}), which can be interpreted as a rotations synchronization problem in the $2$-th power graph, wherein there are edges between any two cameras of $\mathcal{C}$ which are connected via a third vertex in $\mathcal{T}$. The pairwise \textit{measurements} (no longer rotations) of this power graph are the block entries of the block matrix $ \tilde{\mathbf{R}}_{\mathcal{C}\mathcal{T}}\mathbf{\Lambda}_{\mathcal{T}}^{(k-1)^{-1}}\tilde{\mathbf{R}}_{\mathcal{C}\mathcal{T}}^\top$ and the adjacency matrix is taken to be $\mathbf{A}_{\mathcal{CT}}\mathbf{\Delta}_\mathcal{T}^{-1}\mathbf{A}_\mathcal{CT}^\top$. 

The optimization in (\ref{eq:new_update_1}) fits therefore the criteria of the primal-dual method from \cite{Moreira2021RotationGraphs}. Given $\tilde{\mathbf{R}}_{\mathcal{C}\mathcal{T}}\mathbf{\Lambda}_{\mathcal{T}}^{(k-1)^{-1}}\tilde{\mathbf{R}}_{\mathcal{C}\mathcal{T}}^\top$ and $\mathbf{A}_{\mathcal{CT}}\mathbf{\Delta}_\mathcal{T}^{-1}\mathbf{A}_\mathcal{CT}^\top$,  the dual $\mathbf{\Lambda}_\mathcal{C}^{(k)}$ is initialized as the graph degree matrix of the power graph \textit{i.e.}, $\forall c\in{\mathcal{C}}\;\mathbf{\Lambda}_{\mathcal{C}_{i,i}}^{(k)} \leftarrow \sum_j (\mathbf{A}_{\mathcal{CT}}\mathbf{\Delta}_\mathcal{T}^{-1}\mathbf{A}_\mathcal{CT}^\top)_{i,j}$. $\mathbf{R}_\mathcal{C}^{(k)}$ is then obtained by orthogonally projecting the eigenspace spanned by the eigenvectors of $\mathbf{\Lambda}_\mathcal{C}^{(k)}-\tilde{\mathbf{R}}_{\mathcal{C}\mathcal{T}}\mathbf{\Lambda}_{\mathcal{T}}^{(k-1)^{-1}}\tilde{\mathbf{R}}_{\mathcal{C}\mathcal{T}}^\top$ corresponding to the 3 smallest eigenvalues to $\mathrm{SO}(3)^{C}$. $\mathbf{\Lambda}_\mathcal{C}^{(k)}$ is updated according to
\begin{align}
    &\mathbf{U}_i \mathbf{\Sigma}_i \mathbf{V}_i^\top  \stackrel{\mathrm{SVD}}\leftarrow \sum_{j}\left(\tilde{\mathbf{R}}_{\mathcal{C}\mathcal{T}}\mathbf{\Lambda}_{\mathcal{T}}^{(k-1)^{-1}}\tilde{\mathbf{R}}_{\mathcal{C}\mathcal{T}}^\top\right)_{i,j}\mathbf{R}_{\mathcal{C}_j}^{(k)} \\
    &\mathbf{\Lambda}_{\mathcal{C}_i}^{(k)} \leftarrow \mathbf{U}_i\mathbf{\Sigma}_i\mathbf{U}_i^\top.
\end{align}
and the iterations repeat until convergence, upon which (\ref{eq:new_update_2}) is automatically verified. The entire method is laid out in Algorithm (\ref{alg}).

\subsection{Solution for translations}
Minimizing the translations term (\ref{eq:translations_term}) is equivalent to a PGO translation problem with pairwise measurements  
\begin{equation}
    \tilde{\mathbf{t}}_{c,m^{(t)}} := \tau_{c,m^{(t)}} (\tilde{\mathbf{t}}_{c,m^{(t)}} + \mathbf{R}^\ast_c \mathbf{R}_{m^{(t)}}^{\ast^\top} \bar{\mathbf{t}}_{m,m_1}).
\end{equation}
Let $\tilde{\mathbf{T}} \in\mathbb{R}^{3|\mathcal{E}|}$ be a vector containing all the pairwise edges stacked. Define the vector of variables $\mathbf{T} \in\mathbb{R}^{3(C+T)}$
\begin{equation}
    \mathbf{T} := \begin{bmatrix}
        \mathbf{t}_{c_1}^\top & \dots & \mathbf{t}_{c_C}^\top & \mathbf{t}_{m_1^{(1)}}^\top & \dots & \mathbf{t}_{m_1^{(T)}}^\top
    \end{bmatrix}^\top,
\end{equation}
and the incidence block matrix $\mathbf{J} \in \mathbb{R}^{3|\mathcal{E}|\times 3(C+T)}$ such that if $e$ indexes $(c,m^{(t)}) \in \mathcal{E}$ then
\begin{align}
    \mathbf{J}_{e,c} &:= \tau_{c,m} \mathbf{I}_3 \\ 
    \mathbf{J}_{e,t} &:= -\tau_{c,m} \mathbf{R}_c^\ast \mathbf{R}_m^{\ast^\top} \bar{\mathbf{R}}_{m,m_1}.
\end{align}
The translation problem has the least-squares formulation
\begin{equation}
   \min_{\mathbf{T}\in\mathbb{R}^{3(C+T)}} \|\tilde{\mathbf{T}} - \mathbf{J}\mathbf{T}\|_2^2.
   \label{eq:translations_problem}
\end{equation}
Due to the size of the matrices involved, the conjugate gradient method is used to solve (\ref{eq:translations_problem}).

\begin{algorithm}
\caption{Bipartite rotations synchronization}\label{alg}
\begin{algorithmic}[1]
\Require $\{\bar{\mathbf{P}}_{m,m'}\}_{(m,m')\in\mathcal{M}}, \{\tilde{\mathbf{P}}_{c,m^{(t)}}\}_{(c,m^{(t)})\in\mathcal{E}}, \delta, K$
\State $\tilde{\mathbf{R}}_{\mathcal{CT}_{c,t}} \gets \sum_{m:\;m^{(t)}\in N(c)}k_{c,m^{(t)}}\tilde{\mathbf{R}}_{c,m^{(t)}}\mathbf{R}_{m,m_1}$
\State $\mathbf{A}_{\mathcal{CT}_{c,t}} \gets \sum_{m:\;m^{(t)}\in N(c)}k_{c,m^{(t)}}$ \Comment{Adjacency}
\State $\mathbf{\Delta}_\mathcal{T} \gets \mathrm{Diag}\left(\mathbf{A}_{\mathcal{C}\mathcal{T}}^\top\mathbf{1}_C\right)$ \Comment{Degree}
\State $\mathbf{\Lambda}_\mathcal{T} \gets \mathbf{\Delta}_\mathcal{T} \otimes \mathbf{I}_3 $ \Comment{Spectral init.}
\State $\mathbf{\Lambda}_\mathcal{C} \gets \mathrm{Diag}(\mathbf{A}_{\mathcal{C}\mathcal{T}}\mathbf{\Delta}_\mathcal{T}^{-1}\mathbf{A}_{\mathcal{C}\mathcal{T}}^\top\mathbf{1}_C) \otimes \mathbf{I}_3$ 
\Comment{Spectral init.}
\State $k \gets 0$ \Comment{Iteration counter}
\While{$|\lambda_3| > \delta$ and $k < K$}
    \State $\mathbf{L} \gets \mathbf{\Lambda}_\mathcal{C} - \tilde{\mathbf{R}}_{\mathcal{C}\mathcal{T}}\mathbf{\Lambda}_\mathcal{T}^{-1}\tilde{\mathbf{R}}_{\mathcal{C}\mathcal{T}}^\top$
    \State $\mathbf{V},\{\lambda_i\}_{i=1,\dots,3} \gets \mathrm{eigensolver}(\mathbf{L}, 3)$
    \For{$i < C$} \Comment{$\mathcal{C}$ primal update}
        \State $\mathbf{J}_i, \mathbf{\Sigma}_i, \mathbf{H}_i^\top \gets \mathrm{SVD}(\mathbf{V}_{i})$
        \State $\mathbf{R}_{\mathcal{C}_i} \gets \mathbf{J} \mathrm{diag}(1,1,\mathrm{det}(\mathbf{J}_i\mathbf{H}_i^\top))\mathbf{H}_i^\top$
    \EndFor
    \State $\mathbf{S}\gets\tilde{\mathbf{R}}_{\mathcal{C}\mathcal{T}}\mathbf{\Lambda}_\mathcal{T}^{-1}\tilde{\mathbf{R}}_{\mathcal{C}\mathcal{T}}^\top \mathbf{R}_{\mathcal{C}}$ 
    \For{$i < C$} \Comment{$\mathcal{C}$ dual update}
        \State $\mathbf{J}_i, \mathbf{\Sigma}_i, \mathbf{H}_i^\top \gets \mathrm{SVD}(\mathbf{S}_{i})$
        \State $\mathbf{\Lambda}_{\mathcal{C}_{i,i}} \gets \mathbf{J}_i\mathbf{\Sigma}_i\mathbf{J}_i^\top$
    \EndFor
    \State $\mathbf{G} \gets \tilde{\mathbf{R}}_{\mathcal{C}\mathcal{T}}^\top \mathbf{R}_\mathcal{C}$
    \For{$i < T$} \Comment{$\mathcal{T}$ dual update}
        \State $\mathbf{J}_i, \mathbf{\Sigma}_i, \mathbf{H}_i^\top \gets \mathrm{SVD}(\mathbf{G}_{i})$
        \State $\mathbf{\Lambda}_{\mathcal{T}_{i,i}} \gets \mathbf{J}_i\mathbf{\Sigma}_i\mathbf{J}_i^\top$
    \EndFor
    \State $k \gets k + 1$
\EndWhile
\end{algorithmic}
\end{algorithm}

We conclude this section with a note on the feasibility of a streaming version of Algorithm \ref{alg}. At $T+1$, suppose the object moves to a new pose. For all cameras $c\in\mathcal{C}$ with the object in their field-of-view, the relative poses to each of the object nodes, $\tilde{\mathbf{P}}_{c,m^{(T+1)}}$, can be estimated in parallel. Instead of then starting the algorithm again, the initializations in lines 1-5, which would involve increasingly larger matrices for large $T$, can be replaced by  updating the initializations of $\mathbf{L}\in\mathbb{R}^{3C\times 3C}$ and $\tilde{\mathbf{R}}_\mathcal{CT}\mathbf{\Lambda}_\mathcal{T}^{-1}\tilde{\mathbf{R}}_\mathcal{CT}^\top\in\mathbb{R}^{3C\times 3C}$ directly. The optimization problem that follows is only dependent on $C$ and the object poses update is linear in $T$.

\section{EXPERIMENTS}
\label{sec:experiments}
\subsection{Indoor scenes dataset}
\begin{figure*}
    \centering
    \includegraphics[width=\textwidth]{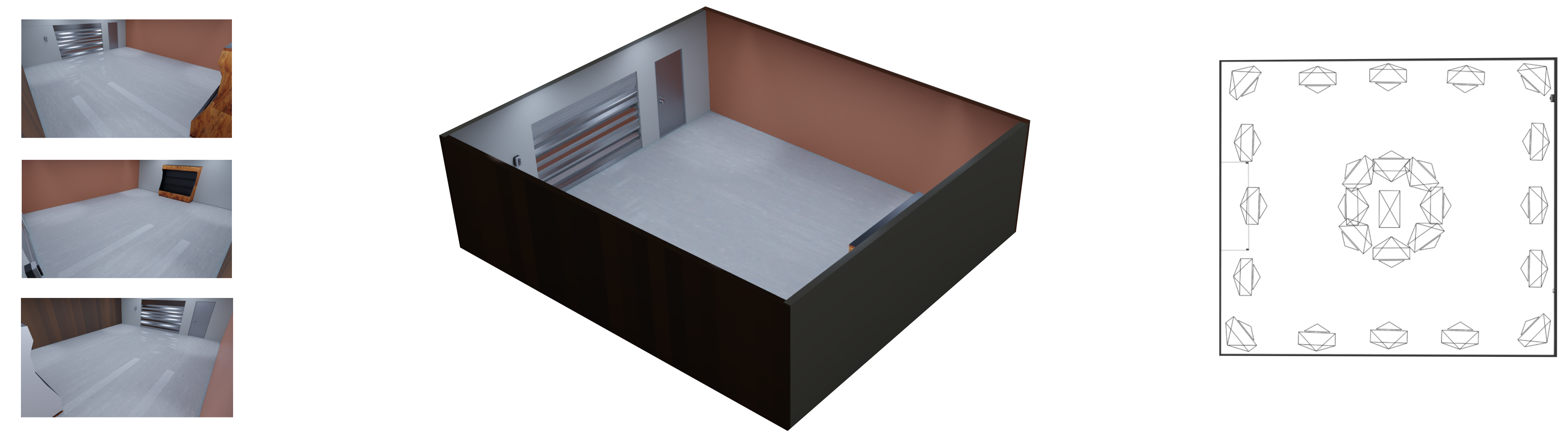}
    \caption{Small room scene: array composed of 25 cameras mounted on the ceiling of a 72m$^2$ room. Left: image examples; Middle: 3D model of the room; Right: top-view of the room with camera locations.}
    \label{fig:small_room}
\end{figure*}

Given that no existing camera network pose estimation dataset is suitable to benchmark the proposed method, we put forward a novel dataset of two camera pose estimation scenes, shown in Figs. 1 and \ref{fig:small_room}. The former consists of a large L-shaped shop with different pieces of furniture and walls blocking the view, whereas the latter is a rectangular room free of occlusions. Images were ray-traced from 3D models of these indoor environments, which were fitted with camera arrays covering them entirely. Each camera has distinct intrinsic ground-truth calibration parameters, that we provide. Given an input value of $T$ time steps, renders from cameras with the chosen object in their field-of-view were obtained procedurally by iteratively placing it $T$ times in random poses, inside the scene. The pose of the object at each time step was sampled uniformly, ensuring no intersections with the environment. An overview of the datasets is shown in Table \ref{table:dataset}. The 3D models, images and the procedural rendering script are available online. 

\subsection{Camera-object transformations and graph construction}
\begin{figure}
    \centering
    \includegraphics[width=0.5\linewidth]{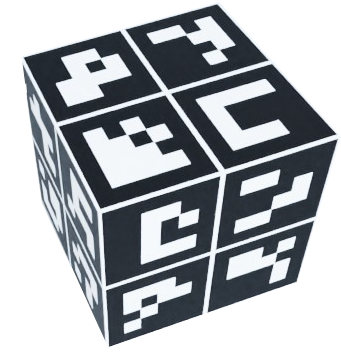}
    \caption{Cube with side 0.575m covered in 24 ArUco markers with side 0.276m, used as the dynamic object in the bipartite pose graph.}
    \label{fig:aruco_cube}
\end{figure}

The object used in all datasets was a cube with side 0.575m, covered in 24 arUco markers, each with side 0.276m, as represented in Fig. \ref{fig:aruco_cube}. Given the set of images obtained as described above, the camera-object edges of the pose graph were obtained via OpenCV's arUco library for corner detection and the implementation of the P4P method provided therein. Further refinement of the camera-marker transformations was done via Levenberg-Marquardt. Finally, edges were filtered according to the average reprojection error. The noise models used for the concentrations $k_{c,m}$ and precisions $\tau_{c,m}$ were derived empirically, based on the detected arUco area in the respective image, as detection accuracy wanes for larger distances and angles.

\subsection{Object calibration}
In order to estimate the relative transformations between any two markers $m$ and $m'$, the cube was fixed in place and 1000 images from a single camera moving around it were captured, from which the pairwise measurements were computed as described above. The estimates $\{\mathbf{P}_m \mathbf{P}_{m'}^{-1}\}_{(m,m')\in\mathcal{M}}$ were obtained by minimizing the NLL (\ref{eq:pgo_neg_log}). Since this graph of the pairwise measurements is bipartite as well, Algorithm \ref{alg} was employed, with the edges reversed \textit{i.e.}, the set of the static cameras $\mathcal{C}$ is the set of 24 static markers and the set of markers of the moving cube $\mathcal{M}\times\mathcal{T}$ becomes the mobile camera. Our algorithm converges in 2 iterations, with an average time per iteration of $0.07s$ in Python.

\begin{table}
\caption{Dataset characteristics}
\label{table:dataset}
\begin{center}
\begin{tabular}{ccccc}
\toprule
Dataset & Area $(m^2)$ & Cams $|\mathcal{C}|$ & Time $|\mathcal{T}|$ & Edges $|\mathcal{E}|$ \\
\hline
SmallRoom50   & \multirow{3}{*}{$72$} & \multirow{3}{*}{$25$} & $50$   & $768$  \\
SmallRoom500  &  &  & $500$  & $8067$   \\
SmallRoom5K   &  &  & $5000$ & $80036$ \\
\hline
LargeShop500  & \multirow{3}{*}{$358$} & \multirow{3}{*}{$342$} & $500$ & $43983$ \\
LargeShop5K   & & & $5000$  & $438906$\\
LargeShop10K  & & & $10000$ & $874700$ \\
\bottomrule
\end{tabular}
\end{center}
\end{table}

\begin{table}
\caption{Pose estimation results: Errors w.r.t. ground-truth}
\label{table:results}
\begin{center}
\begin{tabular}{cccccccc}
\toprule
Dataset & $\mathrm{avg}\;{\delta}_R$ & $\max\delta_R$ & $\mathrm{avg}\;\delta_t$ & $\max\delta_t$ & $t$ (s/it)\\
\hline
SmallRoom50   & $0.54$ & $5.33$ & $0.036$ & $0.285$ & $0.02$ \\
SmallRoom500  & $0.09$ & $0.21$ & $0.008$ & $0.016$ & $0.04$ \\
SmallRoom5K   & $\mathbf{0.07}$ & $\mathbf{0.13}$ & $\mathbf{0.007}$ & $\mathbf{0.012}$ & $\mathbf{0.22}$ \\
\hline
LargeShop500 & $0.09$ & $0.46$ & $0.040$ & $0.093$ &  $0.16$ \\
LargeShop5K  & $0.05$ & $0.15$ & $0.031$ & $0.064$ &  $0.48$ \\
LargeShop10K & $\mathbf{0.04}$ & $\mathbf{0.13}$ & $\mathbf{0.030}$ & $\mathbf{0.064}$ & $\mathbf{0.79}$ \\
\bottomrule
\end{tabular}
\end{center}
\end{table}

\subsection{Camera pose estimation results}
Due to the problem's inherent \textit{gauge symmetry}, given a set of camera pose estimates $\{\mathbf{P}_c\}_{c\in\mathcal{C}}$, for any $\mathbf{H} \in \mathrm{SE}(3)$ the set $\{\mathbf{P}_c \mathbf{H}\}_{c\in\mathcal{C}}$ is also a solution. In order to establish comparisons with the latent camera poses, we consider the equivalence relation $\{\mathbf{P}_c\}_{c\in\mathcal{C}}\sim\{\mathbf{P}'_c\}_{c\in\mathcal{C}}$ $\Leftrightarrow\exists\mathbf{H}\in\mathrm{SE}(3): \mathbf{P}_c = \mathbf{P}'_c\mathbf{H}\;\forall c\in\mathcal{C}$. We define the orbit distance in the quotient manifold $\mathrm{SE}(3)^C / \sim$ as in \cite{Doherty2022PerformanceSLAM} \textit{i.e.},
\begin{align}
    &d_{\mathrm{SE}(3)^C/\sim}\left(\{\mathbf{P}_c\}_{c\in\mathcal{C}}, \{\mathbf{P'}_c\}_{c\in\mathcal{C}}\right) \nonumber \\
    &= \min_{\mathbf{H}\in\mathrm{SE}(3)}d_{\mathrm{SE}(3)^C}\left(\{\mathbf{P}_c\}_{c\in\mathcal{C}}, \{\mathbf{P'}_c\mathbf{H}\}_{c\in\mathcal{C}}\right).
\end{align}
For this choice of gauge, we computed the mean and maximum translation and rotation errors, defined as $\delta_R:=\angle(\bar{\mathbf{R}}_c,\hat{\mathbf{R}}_c)$ in degrees and $\delta_t := \|\bar{\mathbf{t}}_c-\hat{\mathbf{t}}_c\|_2$ in meters, respectively. Results from our Python implementation of Algorithm \ref{alg}, using SciPy sparse matrices and the sparse eigensolver from NumPy (\texttt{eigs}), are shown in Table \ref{table:results}. Convergence was achieved after two iterations in all scenes. The best results are highlighted. 

The different network topologies lead to different graph connectivities. As expected, better results are consistently obtained for better connected graphs. Specifically, the small room scene, devoid of occlusions, lends itself to highly precise camera pose estimation. As anticipated, within both datasets, we note substantial reductions in pose estimation errors with the increase in the number of object poses, without a dramatic rise in Algorithm 1's time per iteration.

\section{CONCLUSIONS}
\label{sec:conclusions}
In this paper, we presented a new approach for pose estimation in large camera networks. Leveraging the well-established duality-based MLE methodology commonly employed in the PGO literature, we laid out a new formulation of the problem as a bipartite pose graph, comprising both camera and object nodes. This allowed us to derive an efficient method to carry out camera pose optimization, with  accuracy improving with additional object poses, but without compromising computational efficiency. In addition, we put forward a new 3D dataset of indoor scenes, for the purpose of camera calibration with and without  objects. We believe that this dataset can serve as a benchmark for future works on object and camera pose estimation. Our algorithm stands as a competitive baseline for the latter.

\section*{Acknowledgement}
This work was supported by  LARSyS funding (DOI: 10.54499/LA/P/0083/2020, 10.54499/UIDP/50009/2020, and 10.54499/UIDB/50009/2020], through Fundação para a Ciência e a Tecnologia. M Marques and J Costeira were also supported by the SmartRetail project [PRR - C645440011-00000062], through IAPMEI - Agência para a Competitividade e Inovação.  


{\small
\bibliographystyle{IEEEtran}
\bibliography{references}
}

\section*{APPENDIX}
\subsection{PGO's negative log-likelihood}
We model the translation measurements through additive isotropic Gaussian noise $\mathbf{\epsilon}_{c,m^{(t)}}$,
\begin{equation}
    \tilde{\mathbf{t}}_{c,m^{(t)}} = \bar{\mathbf{t}}_{c,m^{(t)}} + \mathbf{\epsilon}_{c,m^{(t)}},\quad \mathbf{\epsilon}_{c,m^{(t)}} \sim\mathcal{N}(\mathbf{0}_3,\tau_{c,m^{(t)}}),
\end{equation}
where $\tau_{c,m^{(t)}}=1/\sigma^2_{c,m^{(t)}}$ is the precision of the Gaussian noise and $\bar{\mathbf{t}}_{c,m^{(t)}}$ denotes the latent, or true, translation. From the translations noise model we may write $\mathbf{\epsilon}_{c,m^{(t)}}=\tilde{\mathbf{t}}_{c,m^{(t)}} - \bar{\mathbf{t}}_{c,m^{(t)}} = \tilde{\mathbf{t}}_{c,m^{(t)}} - \bar{\mathbf{t}}_c + \bar{\mathbf{R}}_{c}\bar{\mathbf{R}}_{m^{(t)}}^\top \bar{\mathbf{t}}_{m^{(t)}}$. For all $(c,m^{(t)}) \in \mathcal{E}$ the probability density function (pdf) reads as 
\begin{align}
    &p_\mathcal{N}\left(\mathbf{\epsilon}_{c,m^{(t)}}\right) \propto \nonumber \\
    &\exp\left(-\frac{\tau_{c,m^{(t)}}}{2}\left\| \tilde{\mathbf{t}}_{c,m^{(t)}} - \bar{\mathbf{t}}_c + \bar{\mathbf{R}}_{c}\bar{\mathbf{R}}_{m^{(t)}}^\top \bar{\mathbf{t}}_{m^{(t)}}\right\|_2^2\right).
\end{align}

The rotation measurements are modeled using isotropic Langevin noise $\mathbf{E}_{c,m^{(t)}}$ \cite{Boumal2012Cramer-RaoRotations}
\begin{equation}
    \tilde{\mathbf{R}}_{c,m^{(t)}} = \mathbf{E}_{c,m^{(t)}}\bar{\mathbf{R}}_{c,m^{(t)}},\;\; \mathbf{E}_{c,m^{(t)}} \sim \mathcal{L}(\mathbf{I}_3,k_{c,m^{(t)}}),
\end{equation}
where $k_{c,m^{(t)}} \geq 0$ is the concentration parameter of the isotropic Langevin distribution with density
\begin{equation}
    p_\mathcal{L}(\mathbf{R};\mathbf{S},k)\propto\exp\left(k \mathrm{Tr}\left(\mathbf{S}^\top\mathbf{R}\right)\right)
\end{equation}
and $\bar{\mathbf{R}}_{c,m^{(t)}}$ the latent orientation. The rotations noise model yields $\mathbf{E}_{c,m^{(t)}} = \tilde{\mathbf{R}}_{c,m^{(t)}}\bar{\mathbf{R}}_{c,m^{(t)}}^\top = \tilde{\mathbf{R}}_{c,m^{(t)}}\bar{\mathbf{R}}_{m^{(t)}}\bar{\mathbf{R}}_{c}^\top$. Thus,
\begin{align}
    &p_\mathcal{L}\left(\mathbf{E}_{c,m^{(t)}}\right) \propto  \nonumber \\
    &\exp\left(-k_{c,m^{(t)}}\mathrm{Tr}\left(\tilde{\mathbf{R}}_{c,m^{(t)}}\bar{\mathbf{R}}_{m^{(t)}}\bar{\mathbf{R}}_{c}^\top\right)\right).
\end{align}
Assuming independent measurements, the negative log-likelihood function becomes
\begin{align}
    -\log f = -\log \prod_{c,m^{(t)}\in\mathcal{E}} p\left(\mathbf{\epsilon}_{c,m^{(t)}}\right)p\left(\mathbf{E}_{c,m^{(t)}}\right),
\end{align}
which by taking the logarithm and replacing the summation terms becomes the NLL function in (\ref{eq:pgo_neg_log}).

\subsection{NLL translations term}
Introducing the constraint $\mathbf{t}_{m^{(t)}} = \bar{\mathbf{R}}_{m,m_1} \mathbf{t}_{m_1^{(t)}} + \bar{\mathbf{t}}_{m, m_1}$ in the translations least-squares problem, we have
\begin{align}
    &\sum_{c,m^{(t)}\in\mathcal{E}} \frac{\tau_{c,m^{(t)}}}{2}\|\tilde{\mathbf{t}}_{c,m^{(t)}}-\mathbf{t}_c + \mathbf{R}_{c}\mathbf{R}_{m^{(t)}}^\top \mathbf{t}_{m^{(t)}} \|_2^2 \nonumber \\
    &=\sum_{c,m^{(t)}\in\mathcal{E}}\frac{\tau_{c,m^{(t)}}}{2}\|\tilde{\mathbf{t}}_{c,m^{(t)}}- \nonumber \\ 
    &\quad\quad(\mathbf{t}_c - \mathbf{R}_{c}\mathbf{R}_{m^{(t)}}^\top (\bar{\mathbf{R}}_{m,m_1} \mathbf{t}_{m_0^{(t)}} + \bar{\mathbf{t}}_{m, m_1}))\|_2^2 \nonumber \\
    &=\sum_{c,m^{(t)}\in\mathcal{E}}\frac{\tau_{c,m^{(t)}}}{2}\big\|(\tilde{\mathbf{t}}_{c,m^{(t)}} + \mathbf{R}_c \mathbf{R}_{m^{(t)}}^\top \bar{\mathbf{t}}_{m,m_1}) \nonumber \\ 
    &-(\mathbf{t}_c - \mathbf{R}_{c}\mathbf{R}_{m^{(t)}}^\top \bar{\mathbf{R}}_{m,m_1} \mathbf{t}_{m_1^{(t)}}) \big\|_2^2.
\end{align}

\subsection{NLL rotations term}
Under the constraint $\mathbf{R}_{m^{(t)}} = \bar{\mathbf{R}}_{m,m_1} \mathbf{R}_{m_1^{(t)}}$, the rotations term of the NLL becomes
\begin{align}
    &\sum_{c,m^{(t)}\in\mathcal{E}} - k_{c,m^{(t)}}\mathrm{Tr}(\tilde{\mathbf{R}}_{c,m^{(t)}}\mathbf{R}_{m^{(t)}} \mathbf{R}_c^\top) \nonumber \\
    &=\sum_{c,m^{(t)}\in\mathcal{E}} - k_{c,m^{(t)}}\mathrm{Tr}(\tilde{\mathbf{R}}_{c,m^{(t)}}\bar{\mathbf{R}}_{m,m_1}\mathbf{R}_{m_1^{(t)}} \mathbf{R}_c^\top) \nonumber \\
    &=\sum_{t\in\mathcal{T}}\sum_{c\in\mathcal{C}}\sum_{m^{(t)}\in N(c)} - \mathrm{Tr}\left(k_{c,m^{(t)}}\tilde{\mathbf{R}}_{c,m^{(t)}}\bar{\mathbf{R}}_{m,m_1}\mathbf{R}_{m_1^{(t)}} \mathbf{R}_c^\top\right)  \nonumber \\
    &=\sum_{t\in\mathcal{T}}\sum_{c\in\mathcal{C}} - \mathrm{Tr}\left(\sum_{m^{(t)}\in N(c)}\left(k_{c,m^{(t)}}\tilde{\mathbf{R}}_{c,m^{(t)}}\bar{\mathbf{R}}_{m,m_1}\right) \mathbf{R}_{m_1^{(t)}} \mathbf{R}_c^\top\right) 
\end{align}

\subsection{Rotations synchronization KKT and dual}
For the sake of completeness, we will be deriving the Lagrangian and the dual problem of (\ref{eq:rotation_syncronization_problem}) \cite{Eriksson2018RotationDuality} under the orthogonality constraint $\mathbf{R}_i\mathbf{R}_i^\top = \mathbf{R}_i^\top\mathbf{R}_i = \mathbf{I}_3$, discarding the condition $\mathrm{det}(\mathbf{R}_i) = 1$. Let $\mathbf{\Lambda}\in\mathbb{R}^{3(C+T)\times 3(C+T)}$ be a symmetric block diagonal matrix \textit{i.e.}, 
\begin{equation}
    \mathbf{\Lambda} = \mathrm{BlkDiag}(\mathbf{\Lambda}_{c_1}, \dots, \mathbf{\Lambda}_{c_C},\mathbf{\Lambda}_{m_1^{1}},\dots,\mathbf{\Lambda}_{m_1^{T}}).
\end{equation}
The Lagrangian is
\begin{align}
    L(\mathbf{R},\mathbf{\Lambda}) &= -\mathrm{Tr}\left(\tilde{\mathbf{R}}\mathbf{R}\mathbf{R}^\top\right) + \mathrm{Tr}\left(\mathbf{\Lambda}(\mathbf{R}\mathbf{R}^\top - \mathbf{I}_{3(C+T)})\right) \nonumber\\
    &= \mathrm{Tr}\left(\mathbf{R}^\top(\mathbf{\Lambda}-\tilde{\mathbf{R}})\mathbf{R}\right) - \mathrm{Tr}\left(\mathbf{\Lambda}\right) 
\end{align}
Taking derivatives yields $\nabla_{\mathbf{R}}L = 2(\mathbf{\Lambda}-\tilde{\mathbf{R}})\mathbf{R}$ and $\nabla_{\mathbf{\Lambda}}L = \mathbf{R}\mathbf{R}^\top - \mathbf{I}_{3(C+T)}$ we obtain the first-order KKT condition
\begin{equation}
    \begin{cases}
        (\mathbf{\Lambda}-\tilde{\mathbf{R}})\mathbf{R} = 0, \\
        \mathbf{R}_i\mathbf{R}_i^\top = \mathbf{I}_{3},\quad i\in\mathcal{T}\cup\mathcal{C}.
    \end{cases}
    \label{eq:app:kkt}
\end{equation}
To derive the dual problem, 
\begin{align}
    \inf_{\mathbf{R}} L\left(\mathbf{R},\mathbf{\Lambda}\right) 
    &= \inf_{\mathbf{R}}\left\{\mathbf{R}^\top(\mathbf{\Lambda}-\tilde{\mathbf{R}})\mathbf{R}\right\} - \mathrm{Tr}\left(\mathbf{\Lambda}\right) \nonumber \\
    &= \begin{cases}
    -\mathrm{Tr}\left(\mathbf{\Lambda}\right),\quad \mathbf{\Lambda}-\tilde{\mathbf{R}} \succeq 0 \\
    -\infty, \quad\mathrm{otherwise}.
     \end{cases}
\end{align}
Making the dual feasibility constraint explicit yields the dual SDP problem
\begin{equation}
    \max_{\mathbf{\Lambda}-\tilde{\mathbf{R}}\succeq 0} -\mathrm{Tr}\left(\mathbf{\Lambda}\right) 
\end{equation}
Let $(\mathbf{R}^\ast,\mathbf{\Lambda}^\ast)$ be a primal-dual pair verifying the KKT (\ref{eq:app:kkt}). Then we must have 
\begin{align}
    \mathrm{Tr}\left((\mathbf{\Lambda}^\ast-\tilde{\mathbf{R}})\mathbf{R}^\ast\mathbf{R}^{\ast^\top}\right) = 0,
\end{align}
which means that 
\begin{align}
\mathrm{Tr}\left(\mathbf{\Lambda}^\ast \mathbf{R}^\ast\mathbf{R}^{\ast^\top}\right) = \mathrm{Tr}\left(\mathbf{\Lambda}^\ast\right) = \mathrm{Tr}\left(\tilde{\mathbf{R}}\mathbf{R}^\ast\mathbf{R}^{\ast^\top}\right).
\end{align}
If we further assume that $\mathbf{\Lambda}^\ast-\tilde{\mathbf{R}}\succeq 0$, then the primal-dual pair is feasible and the duality gap is zero \textit{i.e.}, $(\mathbf{R}^\ast,\mathbf{\Lambda}^\ast)$ is optimal.

\subsection{Proof of Theorem \ref{theo:equivalent_problems}}
\begin{proof}
We show that solving the subproblems yields the solution of the original problem. Denote by $(\mathbf{R}_\mathcal{C},\mathbf{\Lambda}_\mathcal{C})$ the primal-dual pair for subproblem (\ref{eq:new_problem_in_c}) and $(\mathbf{R}_\mathcal{T},\mathbf{\Lambda}_\mathcal{T})$ that of subproblem (\ref{eq:new_problem_in_t}). Asterisks denote optimal primal-dual variables of the original problem (\ref{eq:rotation_syncronization_problem}). Assuming strong duality holds. 
\begin{align}
    &\min_{\mathbf{R}_\mathcal{C}\in\mathrm{SO}(3)^C} -\mathrm{Tr}\left(\tilde{\mathbf{R}}_\mathcal{CT}\mathbf{\Lambda}_\mathcal{T}^{\ast^{-1}}\tilde{\mathbf{R}}_\mathcal{CT}^\top \mathbf{R}_\mathcal{C}\mathbf{R}_\mathcal{C}^\top\right) \nonumber \\
    &\stackrel{(a)}=\max_{\mathbf{\Lambda}_\mathcal{C}:\; \mathbf{\Lambda}_\mathcal{C}-\tilde{\mathbf{R}}_\mathcal{CT}\mathbf{\Lambda}_\mathcal{T}^{\ast^{-1}}\tilde{\mathbf{R}}_\mathcal{CT}^\top \succeq 0} -\mathrm{Tr}\left(\mathbf{\Lambda}_\mathcal{C}\right) \nonumber \\
    &\stackrel{(b)}=\mathrm{Tr}\left(\mathbf{\Lambda}_\mathcal{T}^\ast\right)+\max_{\mathbf{\Lambda}_\mathcal{C}:\; \mathbf{\Lambda}-\tilde{\mathbf{R}} \succeq 0, \mathbf{\Lambda}_\mathcal{T}=\mathbf{\Lambda}_\mathcal{T}^\ast}-\mathrm{Tr}\left(\mathbf{\Lambda}\right) \nonumber \\
    &\stackrel{(c)}=\mathrm{Tr}\left(\mathbf{\Lambda}_\mathcal{T}^\ast\right) - \mathrm{Tr}\left(\mathbf{\Lambda}^\ast\right) \nonumber \\
    &=-\mathrm{Tr}(\mathbf{\Lambda}_\mathcal{C}^{\ast}),
\end{align}
(a) from the assumption that strong duality holds; (b) using Schur's complement, the condition $\mathbf{\Lambda}-\tilde{\mathbf{R}}\succeq 0$ is equivalent to $\mathbf{\Lambda}_\mathcal{C}\succeq 0$, $\mathbf{\Lambda}_\mathcal{C}-\tilde{\mathbf{R}}_\mathcal{CT}\mathbf{\Lambda}_\mathcal{T}^{\ast^{-1}}\tilde{\mathbf{R}}_\mathcal{CT}^\top \succeq 0$, $\mathbf{\Lambda}_\mathcal{C}\succeq 0$, $(\mathbf{I}-\mathbf{\Lambda}_\mathcal{C}\mathbf{\Lambda}_\mathcal{C}^{-1})\tilde{\mathbf{R}}_\mathcal{CT}=0$; (c) this is just the dual of (\ref{eq:rotation_syncronization_problem}), with an equality constraint on $\mathbf{\Lambda}_\mathcal{T}$. Similarly, 
\begin{align}
    &\min_{\mathbf{R}_\mathcal{T}\in\mathrm{SO}(3)^T} -\mathrm{Tr}\left(\tilde{\mathbf{R}}_\mathcal{CT}^\top\mathbf{\Lambda}_\mathcal{C}^{\ast^{-1}}\tilde{\mathbf{R}}_\mathcal{CT} \mathbf{R}_\mathcal{T}\mathbf{R}_\mathcal{T}^\top\right) \nonumber \\
    &=\max_{\mathbf{\Lambda}_\mathcal{T}:\; \mathbf{\Lambda}_\mathcal{T}-\tilde{\mathbf{R}}_\mathcal{CT}^\top\mathbf{\Lambda}_\mathcal{C}^{\ast^{-1}}\tilde{\mathbf{R}}_\mathcal{CT} \succeq 0} -\mathrm{Tr}\left(\mathbf{\Lambda}_\mathcal{T}\right) \nonumber \\
    &=\mathrm{Tr}\left(\mathbf{\Lambda}_\mathcal{C}^\ast\right)+\max_{\mathbf{\Lambda}_\mathcal{T}:\; \mathbf{\Lambda}-\tilde{\mathbf{R}} \succeq 0, \mathbf{\Lambda}_\mathcal{C}=\mathbf{\Lambda}_\mathcal{C}^\ast}-\mathrm{Tr}\left(\mathbf{\Lambda}\right) \nonumber \\
    &=\mathrm{Tr}\left(\mathbf{\Lambda}_\mathcal{C}^\ast\right) - \mathrm{Tr}\left(\mathbf{\Lambda}^\ast\right) \nonumber \\
    &=\mathrm{Tr}(\mathbf{\Lambda}_\mathcal{T}^{\ast}).
\end{align}
Hence, solving the dual of subproblem (\ref{eq:new_problem_in_c}) yields $\mathbf{\Lambda}_\mathcal{C}^\ast$ and solving the dual of subproblem (\ref{eq:new_problem_in_t}) produces $\mathbf{\Lambda}_\mathcal{T}^\ast$. Letting $\mathbf{\Lambda}^\ast$ be the block diagonal matrix formed by $\mathbf{\Lambda}_\mathcal{C}^\ast, \mathbf{\Lambda}_\mathcal{T}^\ast$, we can retrieve the optimum of (\ref{eq:rotation_syncronization_problem}) in the kernel of $\mathbf{\Lambda}^\ast-\tilde{\mathbf{R}}$, as per the KKT condition.
\end{proof}

\subsection{Distance in $\mathrm{SE}(3)^C/\sim$}
Consider two sets of poses $\mathbf{P}_c=(\mathbf{R}_c,\mathbf{t}_c)$, $\mathbf{P}'_c=(\mathbf{R}'_c,\mathbf{t}'_c)$ and a gauge $\mathbf{H}=(\mathbf{R}_H,\mathbf{t}_H)$.
\begin{align}
    &\min_{\mathbf{H}\in\mathrm{SE}(3)}d_{\mathrm{SE}(3)^C}\left(\{\mathbf{P}_c\}_{c\in\mathcal{C}}, \{\mathbf{P'}_c\mathbf{H}\}_{c\in\mathcal{C}}\right)^2 \nonumber \\
    &=\min_{\mathbf{H}\in\mathrm{SE}(3)}\sum_{c\in\mathcal{C}}d_{\mathrm{SE}(3)}\left(\mathbf{P}_c, \mathbf{P'}_c\mathbf{H}\right)^2 \nonumber \\
    &=\min_{\mathbf{t}_H\in\mathbb{R}^3}\sum_{c\in\mathcal{C}} \|\mathbf{t}_c - \mathbf{R}'_c\mathbf{t}_H -\mathbf{t}'_c\|_2^2 \nonumber \\
    &- \min_{\mathbf{R}_H\in\mathrm{SO}(3)}\sum_{c\in\mathcal{C}} \mathrm{Tr}\left(\mathbf{R}_c^\top\mathbf{R}_c'\mathbf{R}_H\right).
\end{align}
The gauge translation is thus given by
\begin{equation}
    \mathbf{t}_H = \frac{1}{C}\sum_{c\in\mathcal{C}}\mathbf{R}_c'^\top (\mathbf{t}_c-\mathbf{t}'_c).
\end{equation}
The gauge rotation is the solution of
\begin{equation}
    \min_{\mathbf{R}_H\in\mathrm{SO}(3)}\mathrm{Tr}\left(\left(\sum_{c\in\mathcal{C}} \mathbf{R}_c^\top\mathbf{R}_c'\right)\mathbf{R}_H\right).
\end{equation}
If we denote the singular value decomposition of the sum $\sum_{c\in\mathcal{C}} \mathbf{R}_c^\top\mathbf{R}_c'$ as $\mathbf{U}\mathbf{\Sigma}\mathbf{V}^\top$, then $\mathbf{R}_H = \mathbf{V}\mathbf{U}^\top$.

\end{document}

%% file: fig/measurements.tikz
\tikzset{every picture/.style={line width=0.5pt}} 

\begin{tikzpicture}[x=0.75pt,y=0.75pt,yscale=-1,xscale=1]

\draw   (141.59,19.26) .. controls (141.59,11.06) and (148.25,4.41) .. (156.46,4.41) .. controls (164.67,4.41) and (171.33,11.06) .. (171.33,19.26) .. controls (171.33,27.47) and (164.67,34.12) .. (156.46,34.12) .. controls (148.25,34.12) and (141.59,27.47) .. (141.59,19.26) -- cycle ;
\draw   (267.6,136.99) .. controls (267.6,128.61) and (274.41,121.81) .. (282.8,121.81) .. controls (291.2,121.81) and (298,128.61) .. (298,136.99) .. controls (298,145.37) and (291.2,152.17) .. (282.8,152.17) .. controls (274.41,152.17) and (267.6,145.37) .. (267.6,136.99) -- cycle ;
\draw    (156.46,34.12) -- (156.26,83.34) ;
\draw    (168.62,107.72) -- (199.21,132.99) ;
\draw    (166.35,31.27) -- (214.41,121.81) ;
\draw   (199.21,136.99) .. controls (199.21,128.61) and (206.02,121.81) .. (214.41,121.81) .. controls (222.8,121.81) and (229.61,128.61) .. (229.61,136.99) .. controls (229.61,145.37) and (222.8,152.17) .. (214.41,152.17) .. controls (206.02,152.17) and (199.21,145.37) .. (199.21,136.99) -- cycle ;
\draw    (229.61,136.99) -- (267.6,136.99) ;
\draw   (142.06,98.52) .. controls (142.06,90.14) and (148.86,83.34) .. (157.26,83.34) .. controls (165.65,83.34) and (172.45,90.14) .. (172.45,98.52) .. controls (172.45,106.9) and (165.65,113.7) .. (157.26,113.7) .. controls (148.86,113.7) and (142.06,106.9) .. (142.06,98.52) -- cycle ;
\draw   (298.61,17.47) .. controls (298.61,9.15) and (305.36,2.41) .. (313.68,2.41) .. controls (322.01,2.41) and (328.75,9.15) .. (328.75,17.47) .. controls (328.75,25.8) and (322.01,32.54) .. (313.68,32.54) .. controls (305.36,32.54) and (298.61,25.8) .. (298.61,17.47) -- cycle ;
\draw   (430.32,136.91) .. controls (430.32,128.41) and (437.22,121.51) .. (445.73,121.51) .. controls (454.23,121.51) and (461.13,128.41) .. (461.13,136.91) .. controls (461.13,145.41) and (454.23,152.31) .. (445.73,152.31) .. controls (437.22,152.31) and (430.32,145.41) .. (430.32,136.91) -- cycle ;
\draw   (355.02,137.9) .. controls (355.02,129.4) and (361.91,122.51) .. (370.42,122.51) .. controls (378.92,122.51) and (385.82,129.4) .. (385.82,137.9) .. controls (385.82,146.41) and (378.92,153.3) .. (370.42,153.3) .. controls (361.91,153.3) and (355.02,146.41) .. (355.02,137.9) -- cycle ;
\draw   (299.94,97.87) .. controls (299.94,89.37) and (306.83,82.47) .. (315.34,82.47) .. controls (323.85,82.47) and (330.74,89.37) .. (330.74,97.87) .. controls (330.74,106.37) and (323.85,113.27) .. (315.34,113.27) .. controls (306.83,113.27) and (299.94,106.37) .. (299.94,97.87) -- cycle ;
\draw    (328.15,23.75) -- (383.51,47.6) ;
\draw    (377.53,124.15) -- (392.68,69.47) ;
\draw   (379.07,59.78) .. controls (377.12,58.18) and (376.84,55.3) .. (378.44,53.35) -- (387.79,41.95) .. controls (389.39,40) and (392.27,39.71) .. (394.22,41.31) -- (404.82,50) .. controls (406.77,51.6) and (407.06,54.47) .. (405.46,56.43) -- (396.11,67.83) .. controls (394.51,69.78) and (391.63,70.07) .. (389.68,68.47) -- cycle ;
\draw   (357.64,86.56) .. controls (355.69,84.96) and (355.41,82.09) .. (357.01,80.14) -- (366.36,68.73) .. controls (367.96,66.78) and (370.84,66.49) .. (372.79,68.09) -- (383.39,76.78) .. controls (385.34,78.38) and (385.63,81.26) .. (384.03,83.21) -- (374.68,94.61) .. controls (373.08,96.56) and (370.2,96.85) .. (368.25,95.25) -- cycle ;
\draw    (329.97,91.71) -- (356.1,84.25) ;
\draw    (370.42,122.51) -- (371.25,96.25) ;
\draw    (323.15,28.75) -- (364.37,71.18) ;
\draw   (417.51,66.79) .. controls (415.56,65.19) and (415.27,62.31) .. (416.87,60.36) -- (426.22,48.96) .. controls (427.82,47) and (430.7,46.72) .. (432.65,48.32) -- (443.26,57.01) .. controls (445.21,58.6) and (445.49,61.48) .. (443.89,63.43) -- (434.54,74.84) .. controls (432.94,76.79) and (430.06,77.08) .. (428.11,75.48) -- cycle ;
\draw   (443.59,86.41) .. controls (441.64,84.81) and (441.36,81.94) .. (442.96,79.98) -- (452.31,68.58) .. controls (453.91,66.63) and (456.79,66.34) .. (458.74,67.94) -- (469.34,76.63) .. controls (471.29,78.23) and (471.58,81.11) .. (469.98,83.06) -- (460.63,94.46) .. controls (459.03,96.41) and (456.15,96.7) .. (454.2,95.1) -- cycle ;
\draw    (383.15,128.75) -- (428.11,75.48) ;
\draw    (438,123.5) -- (431.54,76.15) ;
\draw    (457.14,126.27) -- (465.14,88.7) ;
\draw  [dash pattern={on 0.84pt off 2.51pt}]  (384.28,63.5) -- (377.68,72.56) ;
\draw  [dash pattern={on 0.84pt off 2.51pt}]  (447.6,74.0) -- (439.29,68) ;

\draw (150.1,14.8) node [anchor=north west][inner sep=0.75pt]    {$c_{1}$};
\draw (150,93.8) node [anchor=north west][inner sep=0.75pt]    {$c_{2}$};
\draw (207.17,131.8) node [anchor=north west][inner sep=0.75pt]    {$c_{3}$};
\draw (275.8,132) node [anchor=north west][inner sep=0.75pt]    {$c_{4}$};
\draw (183,46) node [anchor=north west][inner sep=0.75pt]  [font=\tiny]  {$\tilde{\mathbf{P}}_{c_{1} ,c_{3}}$};
\draw (234,140.39) node [anchor=north west][inner sep=0.75pt]  [font=\tiny]  {$\tilde{\mathbf{P}}_{c_{3} ,c_{4}}$};
\draw (158,120) node [anchor=north west][inner sep=0.75pt]  [font=\tiny]  {$\tilde{\mathbf{P}}_{c_{2} ,c_{3}}$};
\draw (157.86,69) node [anchor=north west][inner sep=0.75pt]  [font=\tiny]  {$\tilde{\mathbf{P}}_{c_{1} ,c_{2}}$};
\draw (306.89,12) node [anchor=north west][inner sep=0.75pt]    {$c_{1}$};
\draw (308.11,93) node [anchor=north west][inner sep=0.75pt]    {$c_{2}$};
\draw (364,133) node [anchor=north west][inner sep=0.75pt]    {$c_{3}$};
\draw (439,132) node [anchor=north west][inner sep=0.75pt]    {$c_{4}$};
\draw (382,48) node [anchor=north west][inner sep=0.75pt]  [font=\tiny]  {$m_{1}^{( 0)}$};
\draw (360,75) node [anchor=north west][inner sep=0.75pt]  [font=\tiny]  {$m_{2}^{( 0)}$};
\draw (420,55) node [anchor=north west][inner sep=0.75pt]  [font=\tiny]  {$m_{1}^{( 1)}$};
\draw (446,75) node [anchor=north west][inner sep=0.75pt]  [font=\tiny]  {$m_{2}^{( 1)}$};
\draw (340,12) node [anchor=north west][inner sep=0.75pt]  [font=\tiny]  {$\tilde{\mathbf{P}}_{c_{1},m_{1}^{(0)}}$};
\draw (295,40) node [anchor=north west][inner sep=0.75pt]  [font=\tiny]  {$\tilde{\mathbf{P}}_{c_{1},m_{2}^{(0)}}$};
\draw (315,64) node [anchor=north west][inner sep=0.75pt]  [font=\tiny]  {$\tilde{\mathbf{P}}_{c_{2},m_{2}^{(0)}}$};
\draw (330,100) node [anchor=north west][inner sep=0.75pt]  [font=\tiny]  {$\tilde{\mathbf{P}}_{c_{3},m_{2}^{(0)}}$};
\draw (386,128) node [anchor=north west][inner sep=0.75pt]  [font=\tiny]  {$\tilde{\mathbf{P}}_{c_{3},m_{1}^{(1)}}$};
\draw (395,29) node [anchor=north west][inner sep=0.75pt]  [font=\scriptsize]  {$t=0$};
\draw (445,45) node [anchor=north west][inner sep=0.75pt]  [font=\scriptsize]  {$t=1$};
\draw (395,112) node [anchor=north west][inner sep=0.75pt]  [font=\tiny]  {$\tilde{\mathbf{P}}_{c_{4} ,m_{1}^{(1)}}$};

\end{tikzpicture}